\documentclass[letterpaper]{article} 
\usepackage{aaai2026}  
\usepackage{times}  
\usepackage{helvet}  
\usepackage{courier}  
\usepackage[hyphens]{url}  
\usepackage{graphicx} 
\usepackage{natbib}  
\usepackage{caption} 
\usepackage{algorithm}
\usepackage{algorithmic}

\usepackage{kotex}
\usepackage{subcaption}
\usepackage{makecell}
\usepackage{color}
\usepackage{amsmath}
\usepackage{amsfonts}
\usepackage{xcolor}

\definecolor{dark_cyan}{RGB}{0,139,139}
\definecolor{dark_green}{RGB}{0,96,0}
\definecolor{OliveGreen}{rgb}{0.5, 0.5, 0.0}
\definecolor{Fuchsia}{rgb}{1.0, 0.0, 1.0}

\newcommand{\nj}[1]{\textcolor{black}{#1}}
\newcommand{\mj}[1]{\textcolor{black}{#1}}
\newcommand{\hj}[1]{\textcolor{black}{#1}}

\usepackage{newfloat}
\usepackage{listings}
\DeclareCaptionStyle{ruled}{labelfont=normalfont,labelsep=colon,strut=off} 
\floatstyle{ruled}
\newfloat{listing}{tb}{lst}{}
\floatname{listing}{Listing}
\title{Targeted Data Protection for Diffusion Model by Matching Training Trajectory}
\author{
    Hojun Lee\textsuperscript{1}\equalcontrib,
    Mijin Koo\textsuperscript{2}\equalcontrib,
    Yeji Song\textsuperscript{2},
    Nojun Kwak\textsuperscript{2,3}\thanks{Corresponding author.}
}
\affiliations{
    \textsuperscript{1}Xperty Corp. \\ 
    \textsuperscript{2}GSCST, 
    Seoul National University \\ 
    \textsuperscript{3}AIIS, Seoul National University \\
    hojun815@gmail.com,
    \{starmj09, ldynx, nojunk\}@snu.ac.kr
}

\begin{document}

\maketitle

\begin{abstract}

Recent advancements in diffusion models have made fine-tuning text-to-image models for personalization increasingly accessible, but have also raised significant concerns regarding unauthorized data usage and privacy infringement. Current protection methods are limited to passively degrading image quality, failing to achieve stable control. While Targeted Data Protection (TDP) offers a promising paradigm for active redirection toward user-specified target concepts, existing TDP attempts suffer from poor controllability due to snapshot-matching approaches that fail to account for complete learning dynamics. We introduce TAFAP (Trajectory Alignment via Fine-tuning with Adversarial Perturbations), the first method to successfully achieve effective TDP by controlling the entire training trajectory. Unlike snapshot-based methods whose protective influence is easily diluted as training progresses, TAFAP employs trajectory-matching inspired by dataset distillation to enforce persistent, verifiable transformations throughout fine-tuning. We validate our method through extensive experiments, demonstrating the first successful targeted transformation in diffusion models with simultaneous control over both identity and visual patterns. TAFAP significantly outperforms existing TDP attempts, achieving robust redirection toward target concepts while maintaining high image quality. This work enables verifiable safeguards and provides a new framework for controlling and tracing alterations in diffusion model outputs.
\end{abstract}

\section{Introduction}
 \label{sec:intro}
Text-to-image diffusion models~\cite{radford2021learning, Rombach_2022_CVPR, saharia2022photorealistic} have enabled remarkable personalization capabilities through fine-tuning techniques~\cite{gal2023an, ruiz2023dreambooth, kumari2023multi}, enabling custom generation of user-specific concepts. However, these powerful abilities raise significant concerns regarding unauthorized data usage and privacy infringement. The misuse of diffusion models raises significant concerns, including identity theft~\cite{chen2023disenbooth, wang2024simac} for deepfakes and non-consensual content~\cite{westerlund2019emergence}, and copyright infringement~\cite{zhang2023copyright} of intellectual property. These threats pose fundamental challenges to individual privacy, content authenticity, and intellectual property rights, necessitating robust protection mechanisms that can provide comprehensive safeguards against unauthorized usage~\cite{shan2023glaze,shan2023prompt}.

\begin{figure}[t!]
    \centering
    \includegraphics[width=0.98\linewidth]{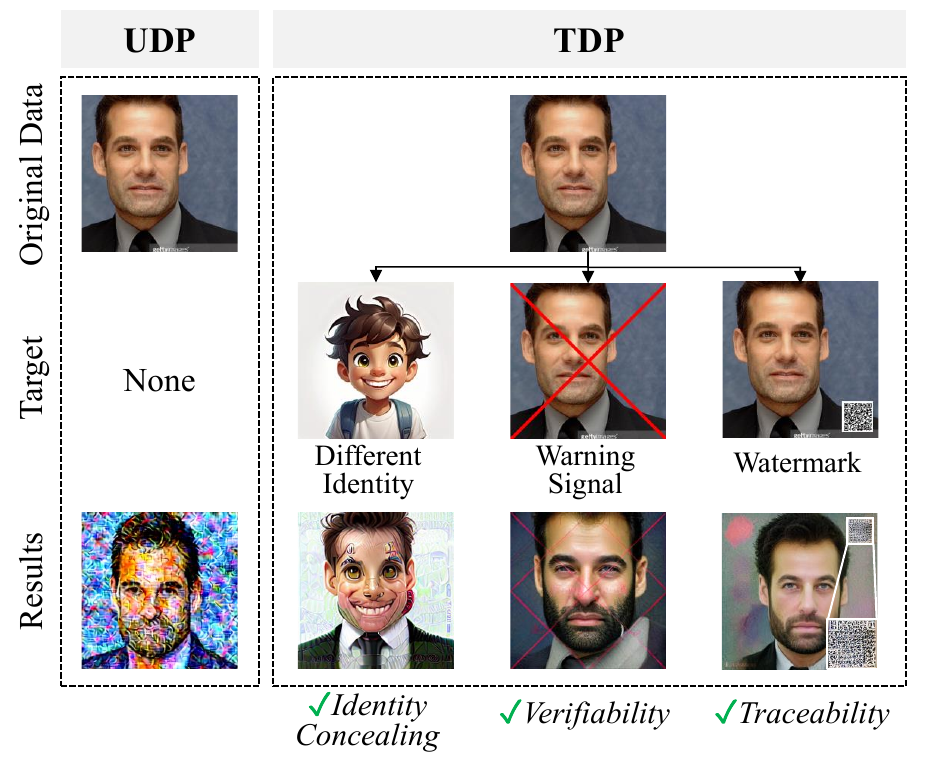}
    \caption{Comparison of Untargeted Data Protection (UDP) and Targeted Data Protection (TDP). UDP passively degrades image quality, while TDP actively redirects toward user-specified targets. \nj{TDP results are from our TAFAP.}}
    \label{fig:teaser_tdp_udp}
\end{figure}

To address these threats, data protection mechanisms must satisfy three critical requirements: 1) \textit{Identity \nj{Concealing}}: The ability to completely anonymize personal features, ensuring models trained on protected data fail to preserve the original person's identifying characteristics. 2) \textit{Verifiability}: The capacity to provide concrete evidence that protection measures are working and to enable clear identification of misuse when it occurs. 3) \textit{Traceability}: The ability to track and attribute unauthorized usage of protected data through distinctive patterns embedded in generated outputs.

\begin{figure*}[t!]
    \centering
    \includegraphics[height=0.36\textheight]{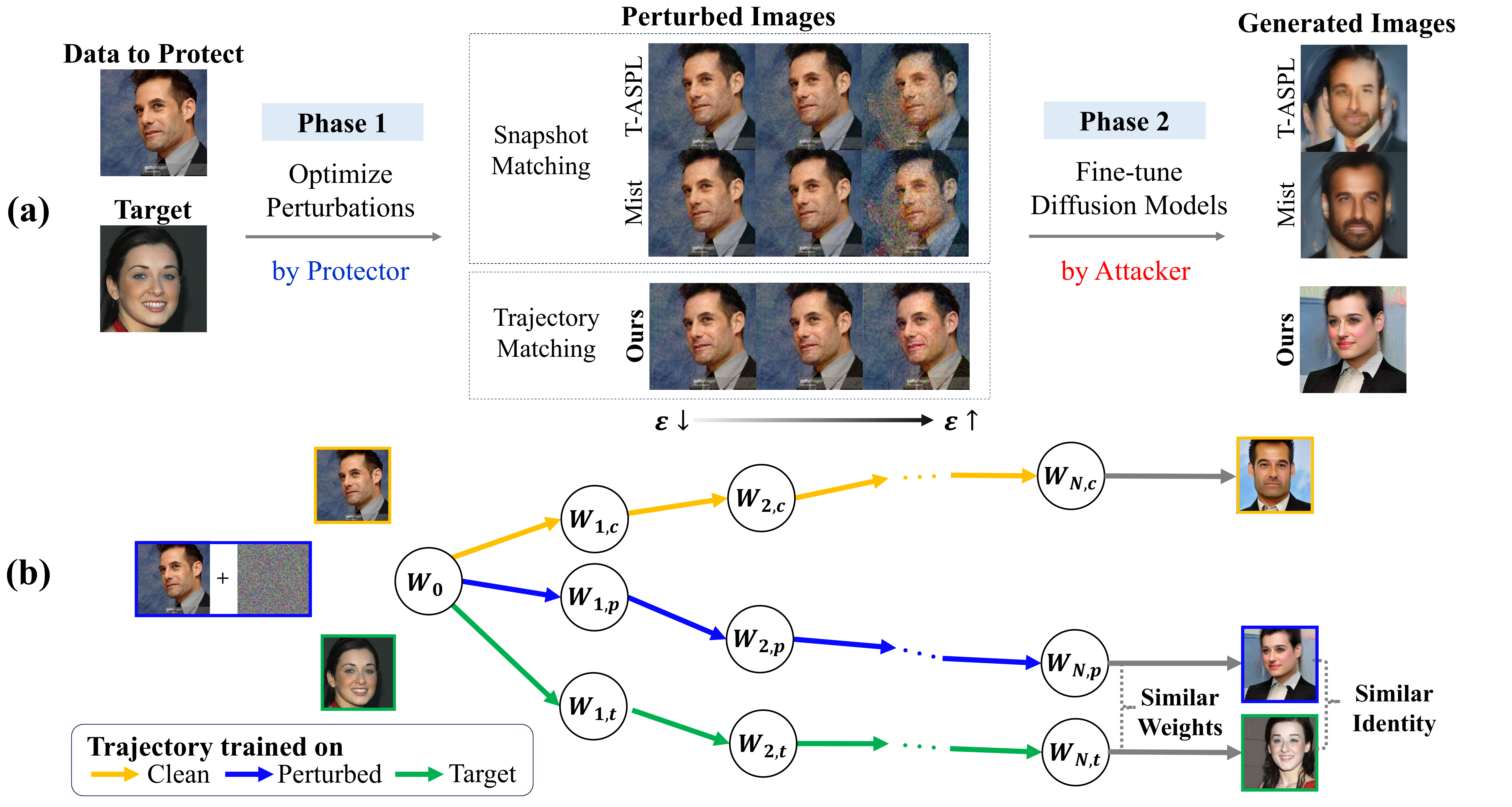}
    \caption{Overview of our method\hj{, TAFAP}. 
    \hj{\textbf{(a) Visual Comparison.} Perturbed images (middle) show that as the perturbation budget ($\delta$) increases, prior works~\cite{liang2023mist,van2023anti} create a perceptual overlap, while TAFAP's modification is semantic. Generated images (right), from the lowest-budget perturbed images shown, demonstrate that only TAFAP successfully generates the target.}
    \textbf{(b) Our goal in Phase 2.} The objective is to align the training trajectory of the protected data with that of the target data, ensuring similar weight updates, thereby enabling the model to learn the \hj{target}
    concept.
    }
    \label{fig:intro}
\end{figure*}

\mj{Current data protection methods primarily rely on adversarial perturbations. Most of these approaches follow} the \textit{Untargeted Data Protection (UDP)} paradigm~\cite{shan2023glaze, liu2024metacloak, zhao2023unlearnable, liang2023adversarial, wang2024simac, ahn2025nearly}, which aims to passively degrade the visual quality of generated images to prevent misuse. However, UDP methods fail to meet the essential requirements outlined above due to their untargeted nature: they often leave residual identity traces~\cite{liu2024disrupting}, provide no verifiable evidence of protection effectiveness, and offer no mechanism for tracing unauthorized use. The core limitation of UDP lies in its passive, destructive approach—it can only disrupt but cannot control what the model learns. This severely constrains its effectiveness in real-world deployments where data owners require robust, accountable, and strategically controllable safeguards.

\mj{These limitations necessitate a fundamental paradigm shift from passive degradation to active, controllable guidance over model outputs.}
\mj{The Targeted Data Protection (TDP) paradigm offers a promising alternative by transforming the protection goal from passive quality degradation to active, controllable redirection toward user-specified target concepts.} \mj{As shown in Fig.~\ref{fig:teaser_tdp_udp},} this targeted approach enables all three critical requirements: complete identity transformation for anonymization, verifiable output patterns for protection confirmation, and distinctive signatures for traceability. 

However, despite its theoretical promise, TDP remains significantly under-explored. \mj{The challenge of achieving precise, semantically meaningful redirection toward specific targets is substantially more complex than simple quality degradation, requiring fine-grained control over the entire learning process. Yet existing TDP attempts~\cite{zheng2025targeted, liang2023mist, van2023anti} directly apply the same \textit{snapshot-matching} approaches originally designed for the much simpler UDP objective. These methods calculate perturbations based only on isolated model states—either initial pre-trained weights or individual fine-tuning checkpoints—an approach fundamentally inadequate for the sophisticated control required by targeted redirection. This methodological inadequacy causes protective influence to be easily diluted as training progresses, resulting in unreliable transformations that fail to achieve semantically meaningful redirection.} 

The \mj{key} insight of our work is that effective targeted protection requires controlling the entire training trajectory, not just individual snapshots. In this paper, we propose \textbf{TAFAP (Trajectory Alignment via Fine-tuning with Adversarial Perturbations)}, the first method to successfully realize Targeted Data Protection by controlling the entire training trajectory of diffusion model fine-tuning. \nj{As illustrated in Fig.~\ref{fig:intro},} unlike existing \textit{snapshot-matching} approaches, TAFAP employs \textit{trajectory-matching} to align the complete optimization path of protected data with that of a target trajectory. This design fundamentally addresses the \mj{shortcomings} of prior methods, which fail to capture long-term learning dynamics and consequently suffer from unstable or diluted protective effects as training progresses. Our \textit{trajectory-matching} approach, inspired by dataset distillation techniques~\cite{cazenavette2022dataset}, enforces persistent, verifiable, and semantically aligned transformations throughout the fine-tuning process, enabling robust and controllable redirection toward user-defined target concepts.

To demonstrate the technical feasibility of the TDP paradigm, we focus on proving its core principle through \textit{identity transformation}—one of the most challenging applications that prior work~\cite{van2023anti} has attempted but failed to achieve reliably. Our extensive experiments demonstrate that TAFAP successfully achieves targeted control over both high-level semantic concepts and low-level visual patterns. This represents the first successful demonstration of genuine targeted transformations in generated images, with simultaneous control over multiple attributes and clear superiority over existing targeted approaches. We anticipate that this work will serve as a cornerstone for future research into more sophisticated and controllable data protection technologies.

The contributions of this work are as follows:
\begin{itemize}
   \item We propose TAFAP, the first method to realize TDP by aligning the entire fine-tuning trajectory, overcoming snapshot-based limitations.
   \item We demonstrate the first successful targeted transformation in diffusion models, achieving simultaneous control over identity and visual patterns.
   \item Our method outperforms existing TDP attempts, offering robust, verifiable, and intent-aligned protection through trajectory-level redirection.
\end{itemize}

\section{Preliminary}

\subsection{Diffusion model and Personalization}
\paragraph{Diffusion model}
Denoising diffusion probabilistic models (DDPMs) \cite{ho2020denoising} define a forward process $q$, gradually adding Gaussian noise to initial real data $x_0$. In contrast, a reverse process involves estimating Gaussian noise at each step. Using the trained reverse process $p_{\theta}(x_{t-1}|x_t)$, one can generate images from the normal distribution $p(x_T) = \mathcal{N}(x_T; 0, I)$. The Latent Diffusion Model (LDM)~\cite{Rombach_2022_CVPR} shifts these processes to the efficient, low-dimensional latent space. 
It consists of two components, an autoencoder and a conditional U-Net~\cite{ronneberger2015u}. The encoder $\mathcal{E}(\cdot)$ of the autoencoder projects a given image $x_0$ to the latent space, yielding $z_0 = \mathcal{E}(x_0)$ while the corresponding decoder $\mathcal{D}(\cdot)$ maps $z_0$ back to the RGB space as $\mathcal{D}(\mathcal{E}(x_0)) \approx x_0$. The conditional U-Net $\varepsilon_{\theta}(\cdot)$ is trained on the latent space, predicting the added Gaussian noise $\varepsilon$ given the noisy latent code $z_t$ at timestep $t$ and the text condition $y$ encoded by the pre-trained CLIP text encoder~\cite{radford2021learning} $\tau(\cdot)$. The training objectives can be formulated as follows:
\begin{small}
\begin{equation} \label{eq:ldm_loss}
   L_{LDM}(\theta| x, y) = \mathbb{E}_{z \sim \mathcal{E}(x), \varepsilon \sim \mathcal {N}(0,1), t}\left [\left \|\varepsilon -\varepsilon _\theta \left (z_t, t, \tau(y)\right )\right \|_2^2\right ]
\end{equation}
\end{small}
\paragraph{Personalization}
As one of the personalization approaches, DreamBooth~\cite{ruiz2023dreambooth} adapts diffusion models to learn a new personalized concept and generate images of that concept in novel contexts. It assigns a unique identifier and class name to represent the new concept, constructing a generic prompt like ``a photo of \textit{sks} [class noun]". Using this prompt, DreamBooth optimizes U-Net or Low Rank Adaptation (LoRA)~\cite{hu2021lora} with the LDM loss (Eq.~\ref{eq:ldm_loss}) to reconstruct the reference images of the concept. Also, the prior preservation loss is adopted to prevent the model from forgetting subjects within the same class as the newly introduced concept. DreamBooth uses the following loss:
\begin{equation} \label{eq:db_loss}
   L_{DB}(\theta) = L_{LDM}(\theta; x_0, c) + \lambda \cdot L_{LDM}(\theta; x_{pr}, c_{pr})
\end{equation}
where $x_0$ and $c$ are the reference images and generic prompts, respectively. The second term represents the prior preservation loss, employing a prior prompt $c_{pr}$ (e.g., ``a photo of [class noun]") and randomly generated images $x_{pr}$ using $c_{pr}$. Hyperparameter $\lambda$ controls importance of the second term.

\subsection{Adversarial attacks} \label{subsec:prelimi_adv_attack}
Adversarial attacks~\cite{szegedy2013intriguing} introduce perturbations to mislead classification models into making incorrect predictions. FGSM~\cite{goodfellow2014explaining} generates adversarial examples $x_{adv}$ by adding small perturbations $\delta$ in the direction of the gradient of the loss function $L$ with respect to the input $x$:
$x_{adv} = x + \varepsilon \cdot \text{sign}(\nabla_x L(\theta, x, y))$. PGD~\cite{madry2017towards} is an iterative extension of FGSM that generates stronger adversarial examples. Recent works extended these techniques to diffusion models~\cite{chen2023advdiffuser, liang2023adversarial}. Adversarial attacks against diffusion models disrupt the model’s ability to predict noise, leading to corrupted outputs or failures in image generation tasks such as editing or personalization. Adversarial attacks can be classified as either untargeted or targeted, depending on whether a specific target is present. In the context of personalization attacks on diffusion models, the goal of an untargeted attack is to degrade personalization by maximizing the DreamBooth loss in Eq.~(\ref{eq:db_loss}), as defined by \cite{van2023anti}.
\begin{equation} \label{eq:untargeted_attack}
    \delta^* = \arg \max_{\delta} L_{DB}(f_\theta(x + \delta), y_{\text{true}}) \quad \text{s.t.} \quad \|\delta\|_p \leq \varepsilon
\end{equation}
On the other hand, the goal of a targeted attack is to mislead personalization toward a specific incorrect target \( y_{\text{target}}\) by minimizing \( L_{\text{DB}} \) .
\begin{equation} \label{eq:targeted_attack}
    \delta^* = \arg \min_{\delta} L_{DB}(f_\theta(x + \delta), y_{\text{target}}) \quad \text{s.t.} \quad \|\delta\|_p \leq \varepsilon
\end{equation}

\subsection{\hj{Foundational Concept: Repurposing Training Trajectory Matching}}
\hj{Our work is built upon the foundational concept of Matching Training Trajectory (MTT) \cite{cazenavette2022dataset}, a technique originally proposed for dataset distillation. The primary goal of MTT is to synthesize a small, efficient dataset by ensuring a model trained on it follows the same parameter update trajectory as one trained on a much larger, real dataset. TAFAP repurposes this concept for a fundamentally different objective: targeted data protection. Instead of generating new data from scratch, we leverage the principle of trajectory alignment to craft imperceptible adversarial perturbations for pre-existing images. The goal is not to compress a dataset, but to misguide the learning process on protected data toward a specific target concept. This strategic adaptation—from a data synthesis tool to an adversarial protection strategy—is a cornerstone of our method's novelty.
}

\begin{figure*}[t!]
    \centering
    \includegraphics[width=0.9\textwidth]{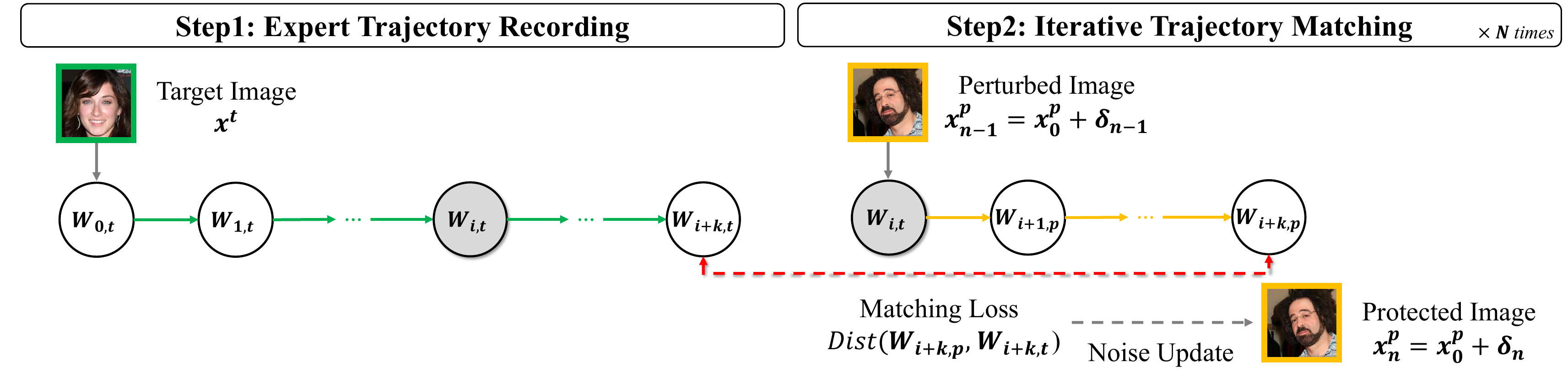}
    \caption{\hj{The two-step process of noise optimization via trajectory matching. \textbf{Step 1}: An expert trajectory is pre-computed by fine-tuning a model on the target data. \textbf{Step 2}: The adversarial noise on the data to protect is iteratively optimized to align the resulting student trajectory with the pre-saved expert trajectory.}
    }
    \label{fig:method_trajectory_matching}
\end{figure*}

\section{Prior Approaches in Data Protection for Text-to-Image Models}  \label{sec:related_work}

Data protection research against fine-tuning with unauthorized data has primarily evolved along two lines: Untargeted Data Protection (UDP) and Targeted Data Protection (TDP). The majority of methods have focused on UDP, a strategy that aims to disrupt the learning process to degrade the overall quality of generated images. While straightforward, this passive approach offers no control over the model's behavior and lacks any mechanism for tracing misuse.

\mj{Recent works have incorporated targeted attack approaches, but with fundamentally different objectives from TDP. Mist~\cite{liang2023mist} and ACE~\cite{zheng2025targeted}} \hj{leverage the mechanism of a targeted attack to enhance UDP effectiveness. By guiding the model toward a predefined chaotic pattern, their objective is not a meaningful transformation, but rather a more severe image degradation. In contrast, T-ASPL~\cite{van2023anti} represents the first attempt at genuine Targeted Data Protection, aiming to redirect identity generation toward specific targets. However, its authors acknowledged that `Targeted methods perform poorly,' failing to achieve meaningful identity redirection while also degrading overall output quality. This outcome led them to focus primarily on untargeted degradation rather than pursuing genuine targeted transformation.}

\mj{The primary obstacle preventing reliable TDP lies in \textit{snapshot-based optimization}. Existing methods, whether UDP-focused (Mist) or TDP-attempting (T-ASPL), rely on myopic optimization that fails to account for complete learning dynamics. Mist calculates perturbations based only on the initial, pre-trained diffusion model—a single snapshot before any fine-tuning occurs. T-ASPL attempts to address this by iteratively calculating perturbations at each fine-tuning step, but because each perturbation only considers the immediate model state without accounting for the full sequence of subsequent weight updates, its influence is easily diluted as training progresses. Considering multiple snapshots is not equivalent to considering the entire training trajectory, and thus cannot guarantee that the model will learn the intended target concept.} 



\hj{In summary, the methodological limitation of existing methods is their reliance on snapshot-based optimization. To overcome this myopic approach and achieve reliable TDP, a new paradigm is needed that can control the entire training trajectory, not just individual snapshots. A more detailed comparison of our approach with prior works is provided in Sec. A of the supplementary material.}

\section{Method} \label{sec:method}
\subsection{Overview} \label{sec:subsec_overview}
In this section, we introduce a method to align the training trajectory of protected data $x^p$ with that of the target data $x^t$ during fine-tuning.
As shown in \hj{Fig.~\ref{fig:intro}b,} when fine-tuning a pre-trained model (e.g., Stable Diffusion), the training trajectories of the protected data and the target data differ. We aim to find appropriate noise \hj{$\delta$} to add to the original data to protect (i.e., \hj{$x^p_0+\delta$}), guiding its trajectory to follow the target trajectory. To achieve this alignment, we measure the distance between the weights for the two trajectories and iteratively adjust the noise added to the protected data. This process aims to minimize the distance between trajectories, effectively guiding the protected data's training trajectory to closely follow that of the target data over time.

\subsection{Adversarial Noise Optimization Process}
While our method's core principle of trajectory matching can be applied to various fine-tuning approaches, our implementation focuses on efficient deployment with minimal computational overhead \cite{hu2021lora}. The key is to track and match the changes in model parameters during fine-tuning, which can be done through various parameter-efficient methods.

\subsubsection{Storing model parameters from Target Data}
To align the training trajectory of the protected data with that of the target data, we store the trajectory of the target data, referred to as the \verb+expert trajectory+. Specifically, we capture the model parameters at each iteration during the target data training process.
These stored parameters are later used to optimize the adversarial noise applied to the protected data. 
Details on memory usage and storage requirements are provided in Sec.~\ref{subsec:experiment_setting}.

\begin{figure*}[t!]
    \centering
    \begin{subfigure}{0.532\textwidth}
        \centering
        \includegraphics[width=0.955\linewidth]{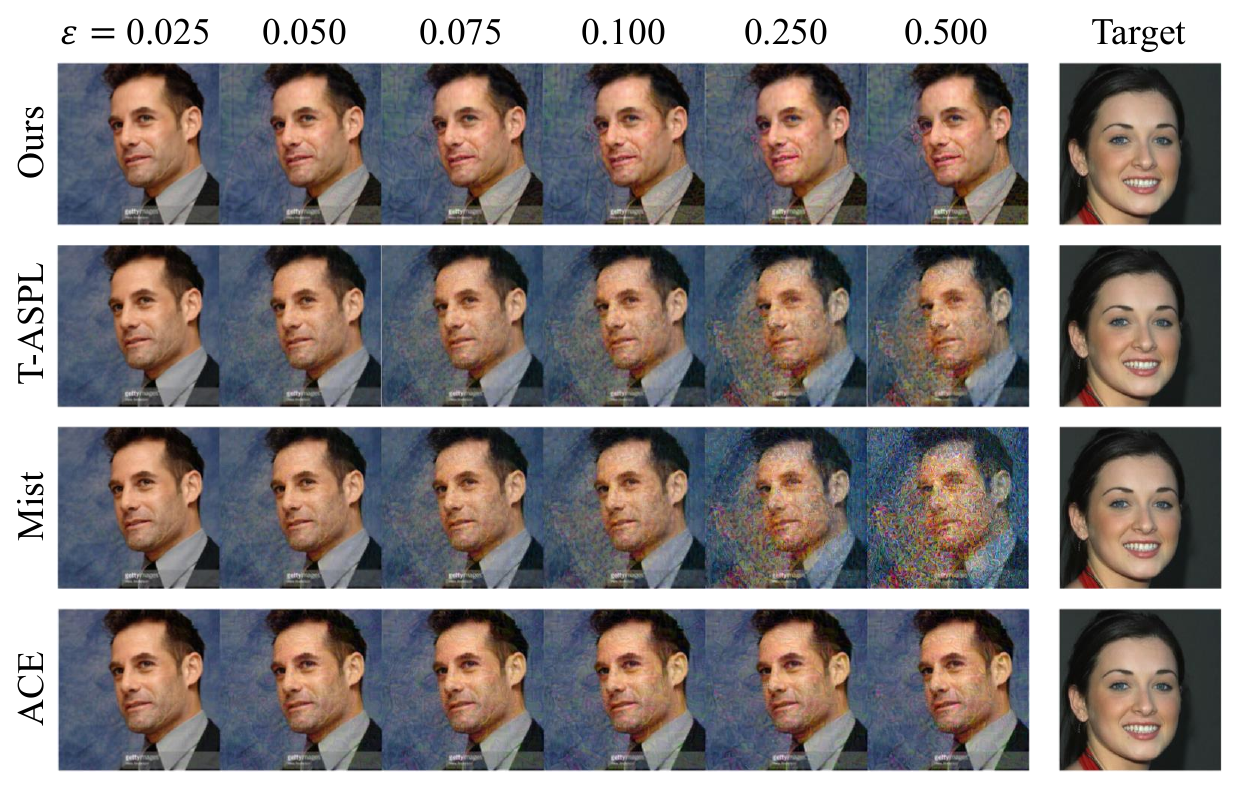}
        \caption{Protected Data with added noise}
        \label{fig:subfig_eps_comparison}
    \end{subfigure}
    \hfill
    \begin{subfigure}{0.448\textwidth}
        \centering
        \includegraphics[width=0.955\linewidth]{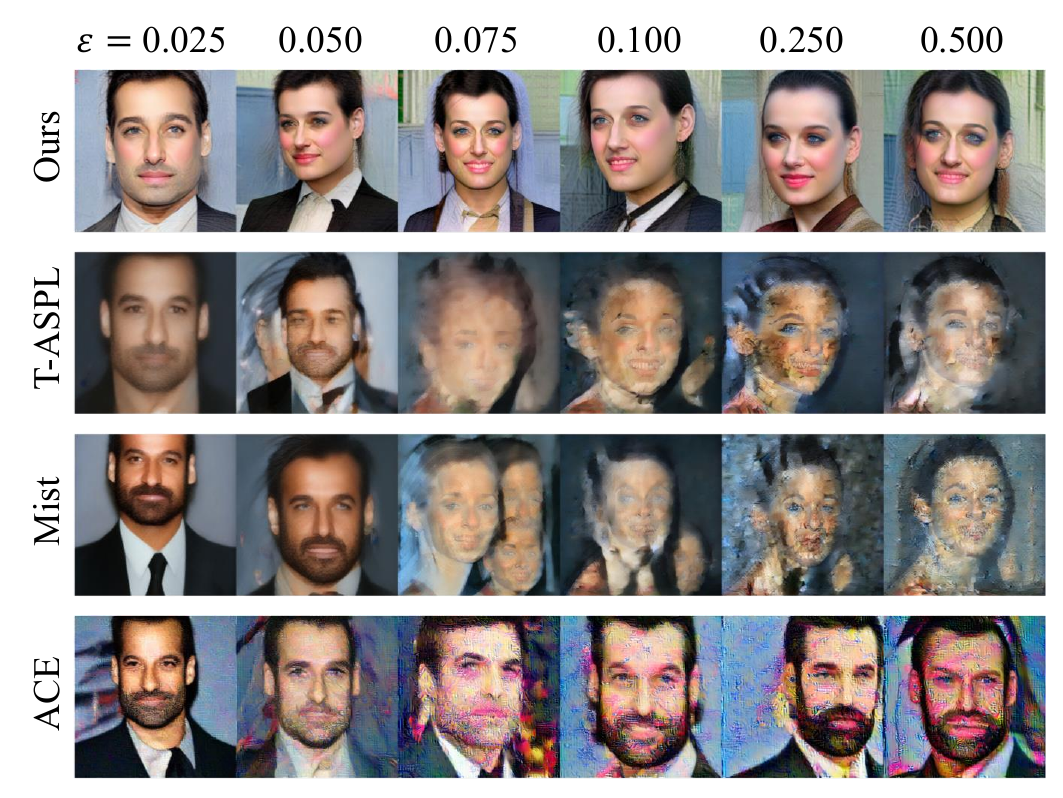}
        \caption{Generation Results}
        \label{fig:subfig_result_comparison}
    \end{subfigure}
    \caption{Comparison with others according to noise budgets $\varepsilon$ \mj{(normalized values)}}
    \label{fig:exp_comparison_with_others_2_fig}
\end{figure*}

\subsubsection{Trajectory matching}
Fig.~\ref{fig:method_trajectory_matching} shows the trajectory matching process. Our goal is to align the training trajectory of the protected data $x^p$ with that of the target data $x^t$ by optimizing the noise $\delta_{N}$ through $N$ iterations. As described in Fig.~\ref{fig:method_trajectory_matching}, we randomly select saved target model parameters and update the noise so that the trajectory of the protected data follows the trajectory of the target data.

At the $n$-th iteration: First, we randomly select the $i$-th target weight $W_{i,t}$. Second, we update the network $k$ times using $x^p_0 + \delta_{n-1}$ to obtain $W_{i+k,p}$. Third, we calculate the distance between $W_{i+k,p}$ and the saved $W_{i+k,t}$. Fourth, we update $\delta_{n-1}$ to minimize this distance. This process is repeated $N$ times.

\paragraph{Objective function} \hj{We use a normalized $L_2$ distance as the objective to ensure consistent loss scaling and guide the protected data’s trajectory toward that of the target:}
\begin{equation} \label{eq:objective_function}
   L = \frac{ \left|\left| W_{i+k,p} - W_{i+k,t} \right|\right|^2_2 }{ \left|\left|W_{i+k,t} - W_{i,t}\right|\right|^2_2  }
\end{equation}



\paragraph{Noise optimization}
To optimize the adversarial noise, we first calculate the gradient of the objective function defined in Eq.~(\ref{eq:objective_function}) with respect to the adversarial noise. Instead of applying the raw gradient values directly, we adopt a sign-based approach, where the gradient of adversarial noise is set to $-1$ for negative values and 1 for positive values, inspired by FGSM~\cite{goodfellow2014explaining} and PGD~\cite{madry2017towards} attacks.
Additionally, a predefined noise budget $\varepsilon$ serves as a constraint, limiting the magnitude of the noise applied to the protected data. Any noise updates that exceed this budget are clipped to prevent excessive perturbations, thereby maintaining the integrity of the protected data while ensuring that the training trajectory aligns closely with the target trajectory.

\section{Experiment}\label{sec:experiment}

\subsection{Experiment Setting}\label{subsec:experiment_setting}
We conducted experiments on the CelebA HQ~\cite{karras2017progressive} and VGGFace2~\cite{cao2018vggface2} datasets. We used Stable Diffusion 1.4~\cite{Rombach_2022_CVPR} as our base model, and for personalization, we employed DreamBooth with Low-Rank Adaptation (LoRA, rank 4). All experiments were performed on a single GeForce RTX 3090 GPU 24GB.


For our setup, both the protected and target data consisted of 12 images each\footnote{To ensure consistency and quality, we manually curated the images, removing those with extreme variations in appearance (e.g., significantly older photos \hj{or facial occlusion}). This process resulted in some identities having 11 images.}, sized at 256$\times$256 pixels. The use of LoRA offered a significant practical advantage, requiring only about 3MB of storage per checkpoint for LoRA weights (6MB including optimizer states) and allowing us to freeze the base model's weights.
\hj{We selected DreamBooth+LoRA as a representative attacker model due to its widespread adoption for low-resource personalization. Nonetheless, our cross-model experiments in Sec.~\ref{subsec:cross-model-generalization} demonstrate the method’s resilience even under attacker model mismatch.}

\mj{The noise budget  $\varepsilon$ was set to 0.05 (12.75/255 in pixel space [0, 255]).}
We used three \verb+expert trajectories+ and optimized the noise \hj{$\delta_N$} through $N=2,000$ iterations.
For DreamBooth with LoRA training, the model was trained for 1,000 iterations. Additional hyperparameters are provided in \hj{Sec. B of the supplementary material.}
\hj{For a fair comparison with Mist~\cite{liang2023mist}, T-ASPL~\cite{van2023anti}, and ACE~\cite{zheng2025targeted}, we utilized only their targeted attack loss components, as these are directly responsible for aligning data with a target concept.}
This setting enables a fair comparison under the Targeted Data Protection (TDP) framework, despite their original focus on untargeted degradation.

\subsection{Qualitative Comparison with other methods}
\paragraph{Comparison of Image Distortions with Respect to Noise Budget} \mj{Fig.~\ref{fig:subfig_eps_comparison} shows the results of adding noise based on noise budgets $\varepsilon$ (normalized to $[0,1]$).} Although the typical goal is to minimize noise to maintain image quality, this experiment is conducted to intentionally observe how different levels of noise impact the visual output. 
The targeted loss functions of Mist~\cite{liang2023mist}, T-ASPL~\cite{van2023anti} and ACE~\cite{zheng2025targeted} are designed to minimize the distance between the latent space of the target data and that of the protected data. As a result, adversarial noise is applied in the image space, often appearing as an afterimage or an overlapping effect. In contrast, our objective function is not designed to reduce the distance between features but to align the training trajectory, which results in fewer visible overlapping effects or afterimages.

\paragraph{Comparison of Data Protection Effectiveness}
Following the previous discussion, Fig.~\ref{fig:subfig_result_comparison} shows the images generated by DreamBooth trained on protected data. While both Mist and T-ASPL, which added noise in a manner that produced afterimages or overlapping effects, failed to generate accurate images, our method was able to generate relatively accurate images following the applied protection.

\subsection{Quantitative Comparison with other methods}
Tab.~\ref{tab:quantitative_comparison} shows the ISM (Identity Score Matching) evaluation results using the ArcFace recognizer~\cite{deng2019arcface}, which measures the similarity between faces. Our method achieved the highest similarity with the target identity while effectively reducing similarity with the protected identity. In contrast, ACE maintained high similarity with the protected identity, and T-ASPL showed low similarity scores with both identities due to face distortion, as visualized in Fig.~\ref{fig:exp_comparison_with_others_2_fig}. 
Note that we evaluated ISM on randomly sampled protect-target pairs from the numerous possible combinations.

Tab.~\ref{tab:qaunt_img_quality} reports image quality metrics including BRISQUE~\cite{mittal2012no}, FDFR, and SER-FIQ. \mj{While such metrics are often used to assess perceptual quality, their interpretation under TDP requires caution. High image quality alone does not imply successful protection unless accompanied by effective semantic redirection, namely the intended identity transformation.} Our method stands out by achieving both high-quality synthesis and effective identity transformation, as confirmed by ISM scores (Tab.~\ref{tab:quantitative_comparison}) and visual comparisons (Fig.~\ref{fig:exp_comparison_with_others_2_fig}).


\begin{table}[t]
    \centering
    \begin{tabular}{ccc}
    \hline
     Method    & ISM w/ data to protect $\downarrow$  & ISM w/ target $\uparrow$ \\ 
    \hline
    No defense  & 0.536  & 0.042 \\
    T-ASPL     & 0.226 & 0.147 \\
    Mist     & 0.368 & 0.108 \\
    ACE     & 0.405 & 0.177 \\
    Ours     & \bf{0.202} & \bf{0.393} \\ 
    \hline
    \end{tabular}
    \caption{Comparison of Identity Score Matching (ISM) across different data protection methods. Lower similarity to the source identity and higher similarity to the target indicate successful targeted transformation.}
    \label{tab:quantitative_comparison}
\end{table}

\begin{table}[t]
    \centering
    \begin{tabular}{cccc}
    \hline
     Method    & BRISQUE $\downarrow$  & FDFR $\downarrow$ & SER-FIQ $\uparrow$\\ 
    \hline
    No defense  & \textbf{1.40} & \textbf{0.000} & \underline{0.78} \\
    T-ASPL      & 23.69 & 0.134 & 0.48 \\
    Mist        & 26.70 & 0.014 & 0.65 \\
    ACE         & 32.74 & 0.019 & 0.67 \\
    Ours        & \underline{11.51} & \textbf{0.000} & \textbf{0.80}\\ 
    \hline
    \end{tabular}
    \caption{Comparison of image quality across protection methods. Higher scores (e.g., lower BRISQUE, higher SER-FIQ) do not directly indicate better protection in TDP. Instead, quality must be considered with identity alignment (Tab.~\ref{tab:quantitative_comparison}) to assess successful redirection toward the target.}
    \label{tab:qaunt_img_quality}
\end{table}


\begin{figure}[t]
    \centering
    \includegraphics[width=0.84\linewidth]{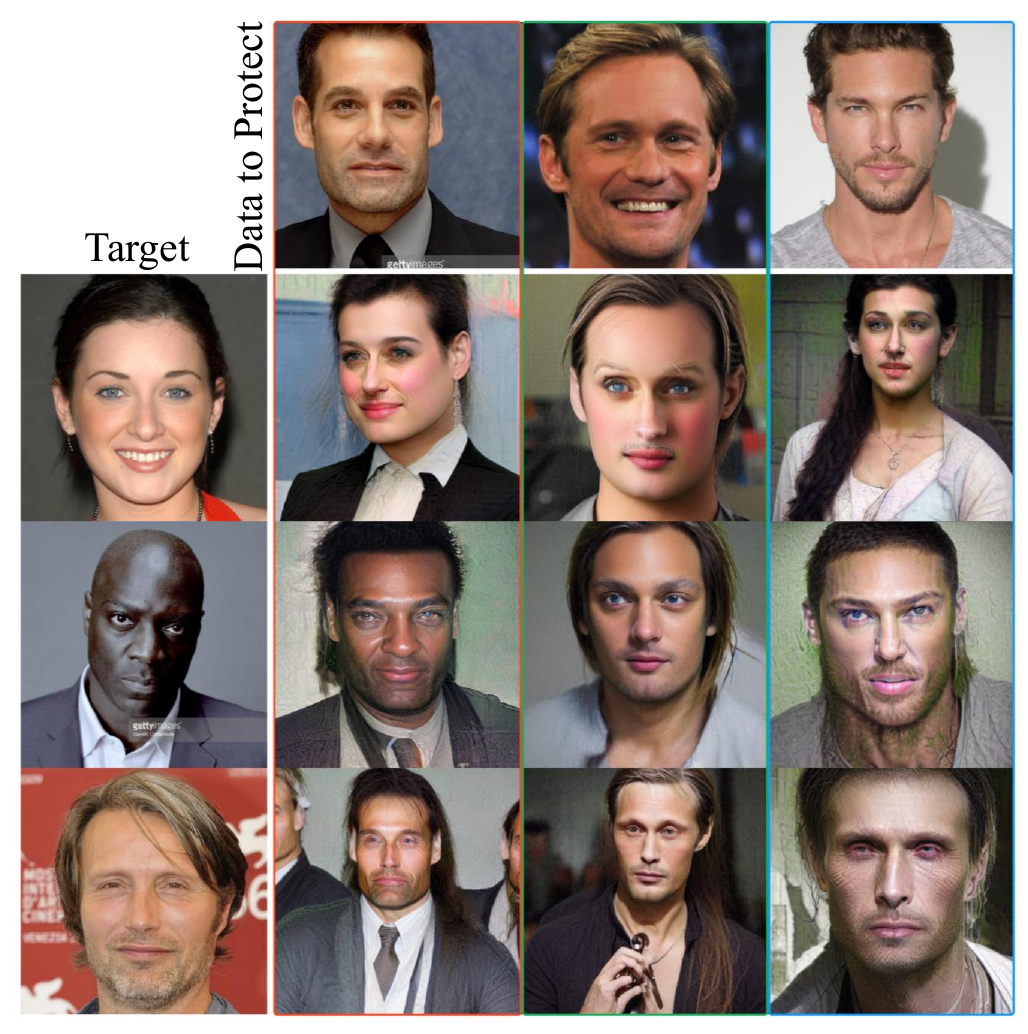}
    \caption{Results of protecting identity from the Personalization method through our TAFAP.}
    \label{fig:exp_ours_various_identity}
\end{figure}

\subsection{Identity Protection Across \hj{Various} Targets} \label{subsec:across_various_target}
Fig.~\ref{fig:exp_ours_various_identity} demonstrates the effectiveness of the proposed method. Despite training on protected data, the model generates images that clearly reflect the target identity. \hj{This confirms our method effectively guides the model's output toward the target.}
Notably, the results vary significantly depending on the chosen target, while using the same protected data. This highlights the flexibility and controllability of our approach, allowing for precise manipulation of the output based on the selected target identity.


\begin{figure}[t]
    \centering
    \includegraphics[width=0.94\linewidth]{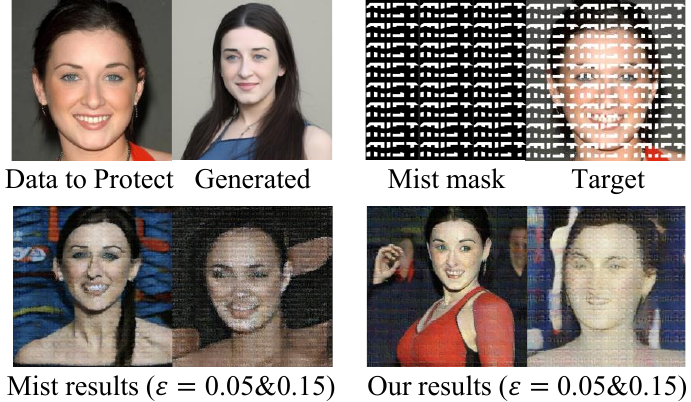}
    \caption{
    \hj{Results when the Mist pattern image is used as the target. Our method successfully redirects generation toward the intended chaotic pattern.}
    }
    \label{fig:exp_mist}
\end{figure}

\subsection{Targeting Chaotic Patterns} 
\label{subsec:pattern_targetting_mist}
\hj{Inspired by the chaotic pattern experiments in Mist~\cite{liang2023mist}, we conducted an experiment to demonstrate the effectiveness of our trajectory alignment method in redirecting the learning process toward complex visual patterns. Consistent adversarial perturbations derived from the Mist mask were applied across all images in our training set, aligning the training trajectory systematically toward the intended chaotic pattern. Fig. \ref{fig:exp_mist} shows results at noise budgets $\varepsilon$ of 0.05 and 0.10. The outcomes demonstrate that our trajectory alignment approach effectively guides the model to generate images clearly reflecting the targeted chaotic patterns. Our method confirms the feasibility of using trajectory-based optimization to achieve controlled and consistent redirection toward complex visual targets.}

\subsection{Effect of Noise Update Iteration}
\begin{figure}[t]
    \centering
    \includegraphics[width=0.93\linewidth]{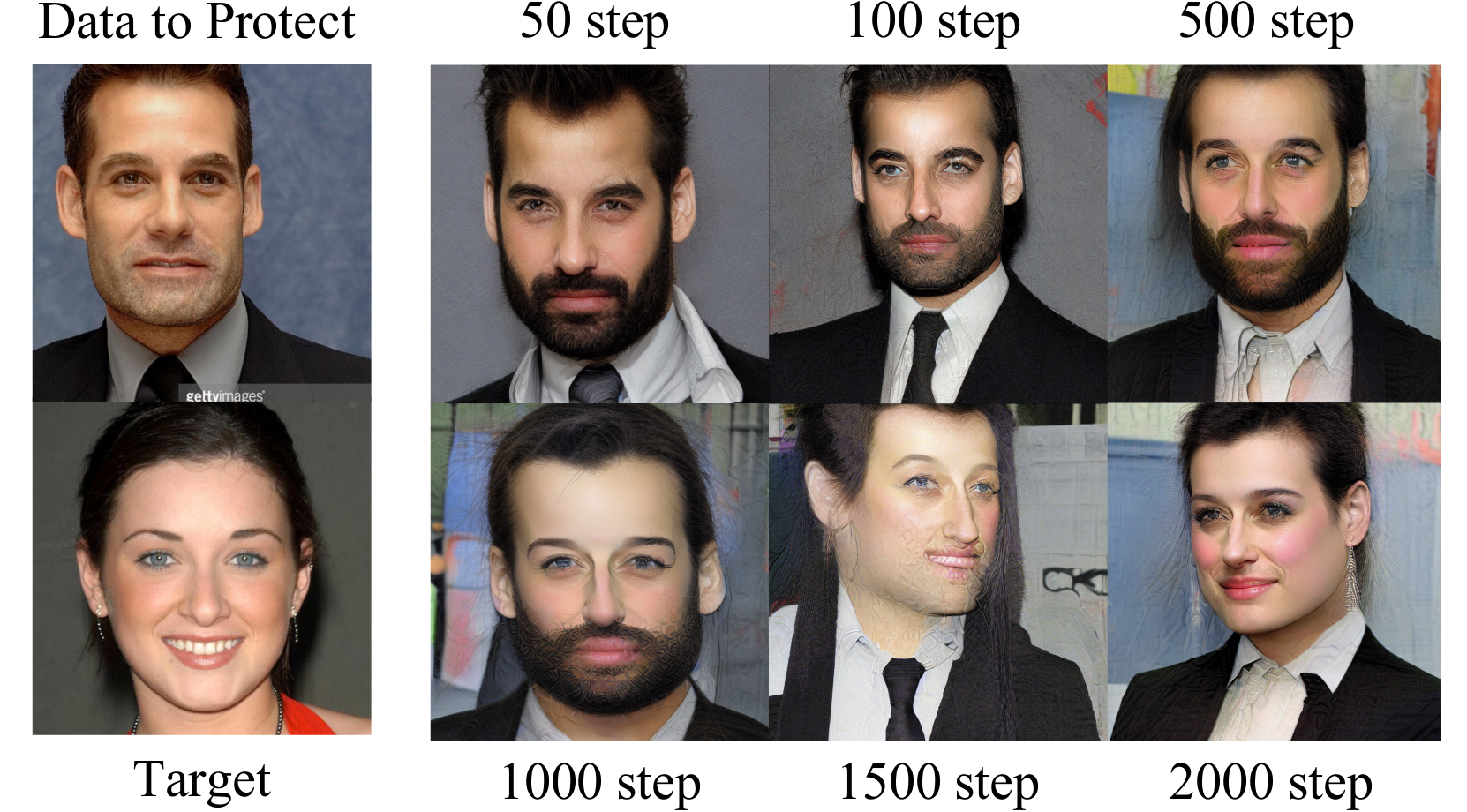}
    \caption{DreamBooth generation results at each step of adversarial noise update.}
    \label{fig:noise_step}
\end{figure}

Fig.~\ref{fig:noise_step} shows the results generated by DreamBooth trained on protected data as we update adversarial noise using our method. As the noise is updated, we can observe that the identity gradually changes, effectively serving the purpose of targeted protection. Notably, this change does not simply affect textures or partial patterns but alters the overall content. This progressive nature of our method suggests the potential for controlling the degree of identity transfer through the number of noise update iterations.

\subsection{Robustness \mj{to} \hj{Image Preprocessing}}
\hj{To emulate common real-world pipelines, we applied three classes of post-processing to the protected images before the attacker’s fine-tune: Gaussian blur $\bigl(k\!\in\!\{3,5,7,9\}\bigr)$, JPEG compression \(\bigl(q\!\in\!\{70,50,30,10\}\bigr)\), and bicubic rescaling–restoration \(\bigl(\downarrow 0.5\times \!\rightarrow\! 1\times\) and \(\uparrow 2\times \!\rightarrow\! 1\times\bigr)\).  
As summarized in Tab.~\ref{tab:gaussian_and_jpeg}, our defense consistently achieves (i) markedly lower similarity to the protected identity and (ii) higher similarity to the target than the No defense baseline across all distortions. Even information-destroying operations such as heavy JPEG compression \((q=10)\) or extreme down-sampling \((\downarrow 0.5\times)\) do not fully negate the redirect effect—the protected-identity score remains well below \(0.536\) while the target-identity score stays above \(0.042\). These findings confirm that the proposed trajectory-based protection retains practical effectiveness under typical, and even aggressive, image-preprocessing conditions.}


\begin{table}[t]
    \centering
    \setlength{\tabcolsep}{3pt}
    \begin{tabular}{ccc}
    \hline
     Method    &  \makecell{ISM with\\data to protect $\downarrow$}  & \makecell{ISM with\\target data $\uparrow$} \\ 
    \hline
     Ours w/o manipulation    & \bf{0.202} & \bf{0.393} \\
    \hline
     Gaussian Blur (kernel=3) & 0.207 &  0.353 \\
     Gaussian Blur (kernel=5) & 0.324 & 0.246 \\
     Gaussian Blur (kernel=7) & 0.484 & 0.169 \\
     Gaussian Blur (kernel=9) & 0.496 & 0.157 \\
    \hline
     JPEG Comp. (quality=70)  & 0.297 & 0.246 \\
     JPEG Comp. (quality=50)  & 0.372 & 0.186 \\
     JPEG Comp. (quality=30)  & 0.426 & 0.171 \\
     JPEG Comp. (quality=10)  & 0.374 & 0.105 \\
    \hline
     Resize $\downarrow 0.5\times \!\rightarrow\! 1\times$ & 0.496 & 0.228 \\
     Resize $\uparrow 2\times \!\rightarrow\! 1\times$    & 0.389 & 0.304 \\
    \hline
     No defense (original data) & 0.536  & 0.042 \\
    \hline
    \end{tabular}
    \caption{Robustness evaluation under preprocessing manipulations on protected images.}
    \label{tab:gaussian_and_jpeg}
\end{table}

\subsection{Cross-Model Generalization} \label{subsec:cross-model-generalization}
\begin{figure}[t!]
    \centering
    \includegraphics[width=0.96\linewidth]{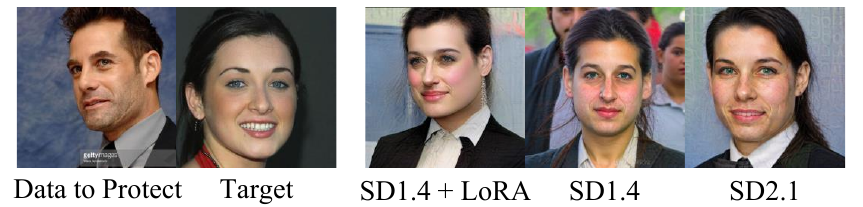}
    \caption{Cross-model effects of our protection method. While optimized with SD1.4+LoRA settings, generated images from different model versions (SD1.4 and SD2.1 without LoRA) still show noticeable deviation from the protected identity, though with varying degrees of target resemblance.}
    \label{fig:cross_model}
\end{figure}

We explore the real-world setting as we do not have complete knowledge about the attacker when training, especially regarding the pre-trained model they may use.
Given the trajectory matching nature of our approach, we expect our method to demonstrate effectiveness across different models. The MTT paper~\cite{cazenavette2022dataset} has shown that trajectory-based synthetic data exhibits strong transferability across different model architectures, suggesting promising potential for our method's cross-model compatibility. Fig.~\ref{fig:cross_model} provides initial evidence for this potential - while our protection is optimized with SD1.4+LoRA, fine-tuning with our protected data remains effective both without LoRA adaptation (SD1.4) and on a different model architecture (SD2.1), showing noticeable deviation from the protected identity in both cases. Exploring this direction could further enhance the practical impact of our approach, particularly in scenarios involving different model versions or architectures.

\section{Conclusion}
\hj{We introduced TAFAP, the first method to achieve Targeted Data Protection through trajectory matching, enabling controlled redirection toward user-specified targets. Unlike snapshot-based approaches that suffer from diluted influence, our trajectory-level control maintains persistent transformations throughout fine-tuning. Using identity transformation as a challenging proof-of-concept, we demonstrated successful protection where prior TDP attempts failed. The observed identity blending in certain cases, rather than being a simple failure, suggests intriguing possibilities for controllable interpolation between protection levels. This work establishes trajectory-based methods as a foundation for proactive and verifiable data protection.}

\section*{Acknowledgments}
The researcher at Xperty was supported by Artificial intelligence industrial convergence cluster development project funded by the Ministry of Science and ICT(MSIT, Korea)\&Gwangju Metropolitan City.
The researchers at Seoul National University were funded by the Korean Government through the grants from NRF (2021R1A2C3006659), IITP (RS-2021-II211343, RS-2025-25442338) and KOCCA (RS-2024-00398320).

\bibliography{aaai2026}
\def\isChecklistMainFile{true}

\clearpage
\appendix

\setcounter{secnumdepth}{2} 

%


\title{Targeted Data Protection for Diffusion Model by Matching Training Trajectory\\--Supplementary Material--}

\renewcommand{\thesection}{\Alph{section}}
\renewcommand{\thefigure}{\Alph{figure}}
\renewcommand{\thetable}{\Alph{table}}

\maketitle


\section{Methodological Comparison of Data Protection Approaches}
\label{appen_sec:comparison_other_method}
\begin{figure*}[t]
    \centering
    \includegraphics[width=0.95\linewidth]{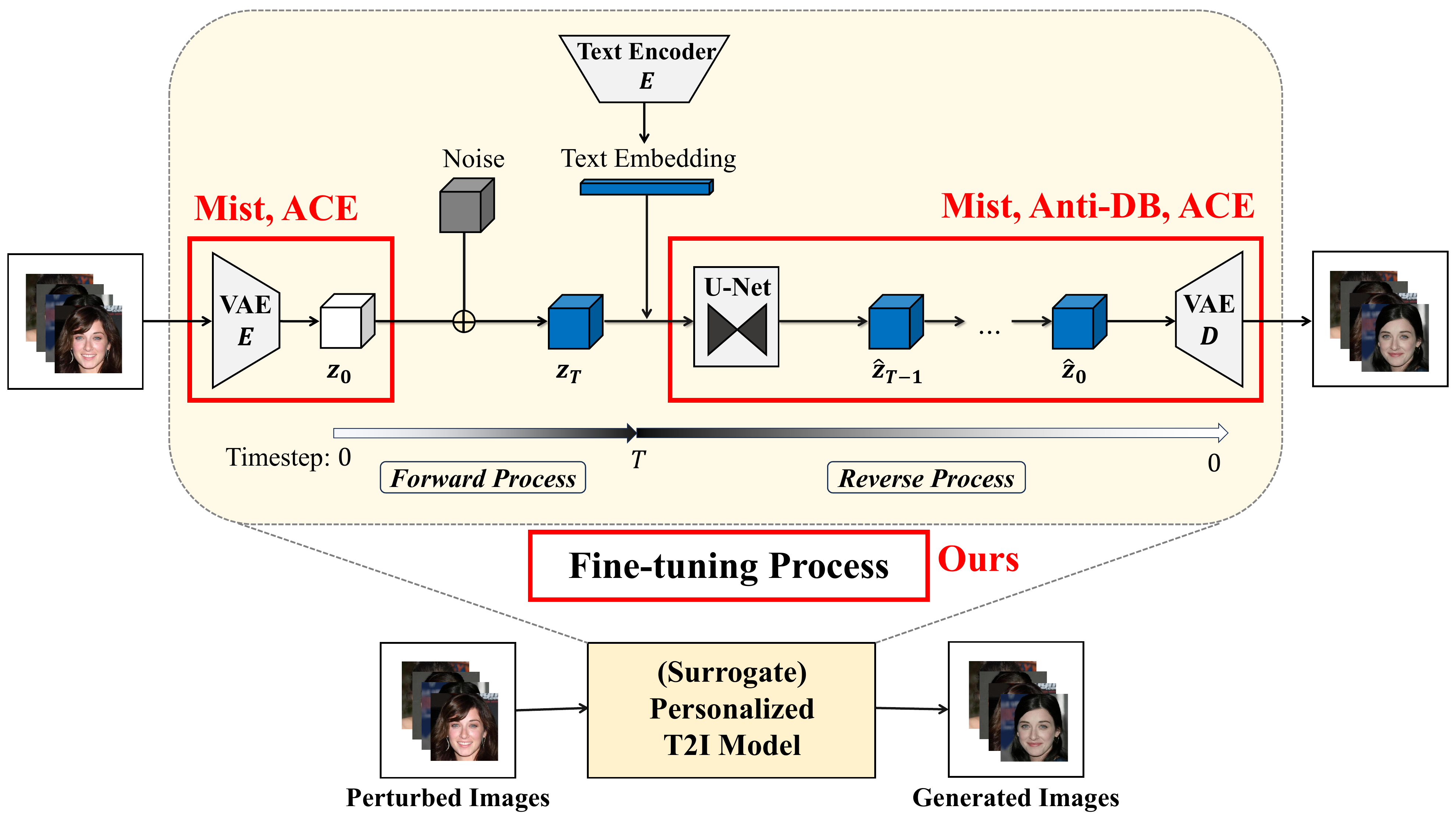}\vspace{-0.5mm}
    \caption{Comparison of data protection methods using targeted attacks, focusing on the attack targets.}
    \label{fig:appen_comparison}
\end{figure*}
In this section, we compare our method to previous targeted approaches that protect data using adversarial perturbations in personalization tasks, focusing on Mist~\cite{liang2023mist} and Anti-DreamBooth~\cite{van2023anti}. As illustrated in Fig.~\ref{fig:appen_comparison}, Mist attacks the VAE encoder and diffusion process to align the encoded $z$ with the targeted representation in the latent space. As an advancement of AdvDM~\cite{liang2023adversarial}, it leverages the semantic loss that disrupts feature extraction in the VAE encoder and the targeting loss that aligns the latent representation with the target.
By aiming to reduce differences in each position of the representation passing through the shallow CNN-based VAE encoder, Mist encounters limitations in protecting high-level information of the entire image. Consequently, as demonstrated in Fig. \hj{4b}, noise is inevitably added in a way that leaves afterimages.

Anti-DreamBooth utilizes a surrogate DreamBooth model to attack the denoising process of the U-Net during fine-tuning, hindering the accurate calculation of predictive noise. While the original objective function of the diffusion model is designed to learn the denoising process from $Z_t$ to $Z_{t-1}$ for the protected data, Anti-DB modifies this approach. Specifically, it generates $Z_t$ from the protected data but derives the target $Z_{t-1}$ from the target data. This modification aims to lead the denoising process to follow the characteristics of the target data rather than those of the protected data. Although this approach effectively disrupts the reverse process, it disregards the assumption that the diffusion model gradually learned Gaussian noise addition process during forward diffusion. As a result, as shown in Fig. \hj{4b}, the generated images are severely distorted, and \hj{Tab. 1} demonstrates that the similarity to the protected data is significantly reduced.

Despite their differences, both Mist and Anti-DB primarily deal with features at specific snapshots rather than addressing how the network should learn.
In contrast, our method targets the fine-tuning process itself, guiding the protected image's fine-tuning trajectory to follow a target training trajectory. We generate noise that alters the trajectory as desired while preserving the process of Gaussian noise prediction inherent to the diffusion model.
This approach offers a more nuanced and controlled method of protection, maintaining the integrity of the underlying diffusion process while effectively redirecting the fine-tuning trajectory.

\section{Additional Implementation Details} \label{appen_sec:implement_detail}
Our implementation involves three main processes: storing the \verb+expert trajectory+ using target data, adding noise to the original data for data protection, and training with the protected data. The following details apply consistently across all three processes. All images were processed using Center Crop. Regarding model precision, LoRA parameters were kept in float32 precision, while all other parameters were in bfloat16 precision. \hj{The text encoder parameters are kept frozen during training, with LoRA fine-tuning applied only to the U-Net component.} For DreamBooth settings, the weight ratio between preservation loss and the loss using the unique identifier (i.e., $sks$) was set to 1:1. We used ``a photo of $sks$ person'' as the instance prompt and ``a photo of person'' as the class prompt. In each of the three processes, LoRA fine-tuning was involved, and for this LoRA fine-tuning, we used a learning rate of 1e-4 without any learning rate scheduler or warmup.  The batch size was set to 1, and no gradient accumulation was performed.

\mj{For our trajectory matching optimization in Sec. 4.2, we set the maximum adversarial training steps to $N=2000$. The PGD optimization uses 6 steps with step size $\alpha=0.001$ and noise budget $\varepsilon=0.05$. The expert trajectory storage utilizes 3 target LoRA models with trajectory snapshots captured every iteration. 
\hj{We employ the normalized trajectory matching loss (Eq. 5) with network update steps $k$=8, which we found to provide optimal balance between trajectory alignment effectiveness and computational efficiency.}
The trajectory matching process spans from iteration 1 to 600 of the expert training. All experiments use random seed 0 for reproducibility.}

\section{On the Numerical Stability of Bi-Level Optimization}
As described in Sec. 4, our TAFAP method formulates trajectory matching as a bi-level optimization problem. The outer loop optimizes adversarial noise $\delta$ by backpropagating through the entire inner loop simulation of the fine-tuning process.

This unrolled inner loop creates a computationally deep graph spanning multiple gradient computation steps. A well-known challenge in such deep computational graphs is the vanishing gradient problem \cite{hanin2018neural,pascanu2013difficulty}, where gradients can exponentially decay during backpropagation through many layers, potentially leading to numerical underflow in floating-point arithmetic.

To ensure robust gradient flow back to the input images, we introduce a constant scaling factor $\lambda = 100$ applied to the trajectory matching loss in Eq. 5. Crucially, this is not a tunable hyperparameter that affects the protection outcome. As detailed in our PGD-based noise optimization (Sec. 4.2), the final noise update is determined solely by the sign of the $\text{sign}(\nabla_{x^p} L)$ with respect to the perturbed images, not by its magnitude.

Therefore, the loss scaling serves solely as a numerical safeguard to prevent gradient underflow while preserving the gradient direction—the only component used for noise updates. This ensures the integrity of our trajectory matching optimization without influencing the method's behavior.

\section{Experimental Setup}
\paragraph{Experimental Setups on Baselines} \mj{For fair comparison with baseline methods, we carefully adapted their experimental settings to align with our TDP evaluation framework. We compared against T-ASPL (Van Le et al. 2023), Mist (Liang and Wu 2023), and ACE (Zheng, Liang, and Wu 2025), using identical target images and PGD attack configurations across all methods. While all three baselines employ targeted attack frameworks, their original objectives differ significantly from our TDP goals. T-ASPL was originally designed for targeted identity redirection and used different identity images as targets, consistent with TDP principles. However, Mist and ACE were primarily developed for UDP purposes, originally using chaotic pattern images as targets to maximize image degradation rather than achieve meaningful identity transformation. To enable fair evaluation under the TDP paradigm, we utilized only the targeted attack loss components of Mist and ACE, replacing their original chaotic targets with actual identity images. This modification allows us to assess their potential for controlled identity redirection while maintaining their core algorithmic approaches. All methods use the same noise budget $\varepsilon = 0.05$, identical personalization settings (DreamBooth + LoRA), and consistent evaluation metrics to ensure direct comparability of protection effectiveness.}

\paragraph{Evaluation Metrics} \mj{Following \hj{previous methods}~\cite{van2023anti, wang2024simac, zheng2025targeted}, we utilize Identity Score Matching (ISM) and Face Detection Failure Rate (FDFR) as our primary evaluation metrics for face-based protection assessment, complemented by image quality metrics. FDFR employs RetinaFace~\cite{deng2020retinaface} to evaluate face detectability in generated images, calculated as the percentage of images where face detection fails, \hj{where lower values indicate better face detection}. ISM quantifies identity preservation by extracting face embeddings via ArcFace~\cite{deng2019arcface} and computing cosine similarity between generated and reference face representations\hj{, where higher values indicate greater facial similarity}. For TDP evaluation, we assess ISM in two distinct aspects: \hj{the similarity between generated images and the original identity, and the similarity between generated images and the target identity.}
In addition, the image quality is quantified by BRISQUE~\cite{mittal2012no}, which assesses perceptual image quality without requiring reference images using natural scene statistics, with scores ranging from 0 to 100 where lower values indicate better quality. The quality of detected facial images is measured through SER-FIQ~\cite{terhorst2020ser}, a specialized face image quality assessment with scores ranging from 0 to 1 where higher values indicate better face-specific image quality, making it more appropriate than general image quality metrics for face-focused protection applications.}

\paragraph{Interpretation Guidelines for TDP}
\mj{The evaluation of Targeted Data Protection (TDP) methods requires fundamentally different interpretation criteria compared to Untargeted Data Protection (UDP) approaches. UDP methods aim to completely disrupt face generation and expect high FDFR through face detection failure. In contrast, our TDP approach seeks controlled identity transformation while preserving face structure. Therefore, TDP requires low FDFR to maintain face detectability. For identity metrics, \hj{while }UDP methods \hj{aim for}
uniformly low ISM by disrupting facial identification\hj{,} TDP requires dual evaluation with opposing goals: \hj{a lower ISM between generated images and the original identity to protect indicates better identity concealing, while a higher ISM between generated images and target identity indicates better targeted redirection.}
\hj{Consequently, a} successful TDP method must simultaneously achieve three characteristics: (1) effective identity concealing (low ISM \hj{with data to protect}), (2) successful target redirection (high ISM \hj{with} target data), and (3) preserved face detectability (low FDFR). Importantly, high image quality scores alone do not indicate protection effectiveness. They must be accompanied by successful identity transformation metrics. The critical distinction \hj{—measuring success by transformation quality and controllability rather than disruption degree—}
establishes a fundamentally different evaluation paradigm from traditional protection approaches.}


\end{document}